\documentclass[letterpaper, 10 pt, conference]{ieeeconf}  

\IEEEoverridecommandlockouts                              

\usepackage{graphics} 
\usepackage{epsfig} 
\usepackage{times} 
\usepackage{amsmath} 
\usepackage{amssymb}  

\usepackage{cancel}
\usepackage{multirow}
\usepackage{booktabs}
\usepackage{balance}

\title{\LARGE \bf Gait Sequence Upsampling using Diffusion Models \\ for Single LiDAR Sensors}

\author{Jeongho Ahn$^{1}$, Kazuto Nakashima$^{2}$, Koki Yoshino$^{1}$, Yumi Iwashita$^{3}$ and Ryo Kurazume$^{2}$%
\thanks{This work was supported in part by the Japan Science and Technology Agency (JST) SPRING under Grant JPMJSP2136, and in part by the Japan Society for the Promotion of Science (JSPS) KAKENHI under Grant JP20H00230.}%
\thanks{Jeongho Ahn and Koki Yoshino are with Graduate School of Information Science and Electrical Engineering, Kyushu University, Japan
        {\tt\small \{ahn, yoshino\}@irvs.ait.kyushu-u.ac.jp}}%
\thanks{Kazuto Nakashima and Ryo Kurazume are with Faculty of Information Science and Electrical Engineering, Kyushu University, Japan
        {\tt\small \{k\_nakashima, kurazume\}@ait.kyushu-u.ac.jp}}%
\thanks{Yumi Iwashita is with Jet Propulsion Laboratory, California Institute of Technology, USA
        {\tt\small yumi.iwashita@jpl.nasa.gov}}%
}

\begin{document}

\maketitle
\thispagestyle{empty}
\pagestyle{empty}

\begin{abstract}


%
Recently, 3D LiDAR has emerged as a promising technique in the field of gait-based person identification, serving as an alternative to traditional RGB cameras, due to its robustness under varying lighting conditions and its ability to capture 3D geometric information.
However, long capture distances or the use of low-cost LiDAR sensors often result in sparse human point clouds, leading to a decline in identification performance.
To address these challenges, we propose a sparse-to-dense upsampling model for pedestrian point clouds in LiDAR-based gait recognition, named LidarGSU, which is designed to improve the generalization capability of existing identification models.
Our method utilizes diffusion probabilistic models (DPMs), which have shown high fidelity in generative tasks such as image completion.
In this work, we leverage DPMs on sparse sequential pedestrian point clouds as conditional masks in a video-to-video translation approach, applied in an inpainting manner.
We conducted extensive experiments on the \textit{SUSTeck1K} dataset to evaluate the generative quality and recognition performance of the proposed method.
Furthermore, we demonstrate the applicability of our upsampling model using a real-world dataset, captured with a low-resolution sensor across varying measurement distances.

\end{abstract}

\section{INTRODUCTION}


%
Gait recognition is pivotal in the field of person identification.
Unlike other biometric modalities, such as facial recognition, retinal scans, or fingerprints, gait offers distinct advantages, including the ability to identify individuals from a distance without requiring their cooperation, and it stands out for being non-intrusive.
These unique physical characteristics make gait recognition particularly well-suited for applications in security systems and criminal investigations.
\begin{figure}[t]
    \centering
    \includegraphics[width=0.48\textwidth]{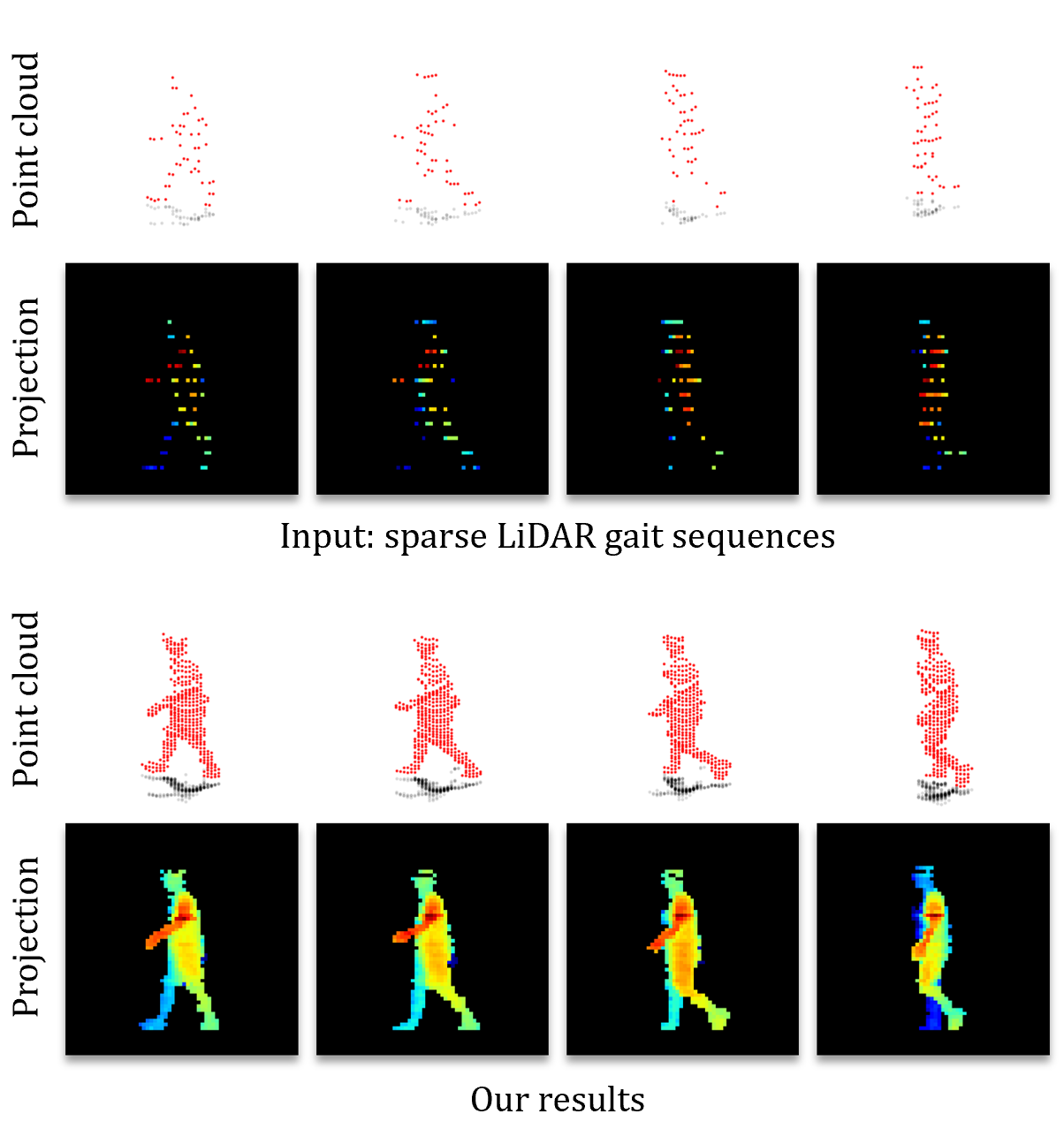}
    \caption{Upsampled results using our models. We present sparse LiDAR gait sequence data as inputs (top two rows) alongside the corresponding outputs (bottom two rows), represented in both 3D point cloud sequences (rows 1 and 3) and 2D depth videos (rows 2 and 4).}
    \label{fig:overview}
\end{figure}
%

%
In recent years, light detection and ranging (LiDAR) has emerged as an alternative technique in the field of gait recognition, playing a critical role in mobile robotics and self-driving cars owing to its ability to capture 3D point clouds of surrounding obstacles and geometries by emitting laser beams.
Previous studies on LiDAR-based gait recognition \cite{lidargait} have reported that it outperforms traditional RGB cameras, owing to its robustness against varying lighting conditions and capacity to provide accurate geometric information, making it more suitable for security applications.
However, when deploying LiDAR sensors in identification systems, the density of human point clouds is highly sensitive to measurement distance and hardware specifications, especially when using low-resolution sensors.
This sensitivity often results in significant degradation in identification performance due to the sparse or incomplete data of human shapes.

%
In this paper, we present a LiDAR-based gait sequence upsampling model for sparse pedestrian point cloud data, named LidarGSU (as shown in Fig. \ref{fig:overview}), with the aim of enhancing the generalization capability of existing identification models.
Our method utilizes diffusion probabilistic models (DPMs) \cite{ddpm}, which have shown high fidelity in generative tasks, including image completion.
Specifically, we treat the missing points within gait shapes using a distance-independent inpainting strategy by projecting pedestrian point clouds into a 3D Euclidean space, feeding them into a diffusion-based architecture with corresponding conditional masks.
Furthermore, to ensure consistency in time-sequential gait appearances, we employed a video-based noise prediction model \cite{vdm} during the denoising process.
To the best of our knowledge, this is the first study to address LiDAR data upsampling for gait recognition.
In our experiments, we demonstrated the effectiveness of our model on two datasets: the \textit{SUSTeck1K} dataset \cite{lidargait} and Ahn \textit{et al.}'s dataset \cite{2v-gait-v2}, both of which evaluate both generative quality and improvements in identification performance.

%
The contributions of this study can be summarized as follows:
\begin{itemize}
    \item{We present a LiDAR upsampling method based on conditional diffusion models that utilizes a distance-independent inpainting approach to enhance the generalization capability of existing LiDAR-based recognition models.}
    \item{By employing a video-based noise prediction technique, our diffusion model ensures consistency in the sequential pedestrian gait shapes. In addition, we used a continuous time schedule for fast and efficient LiDAR upsampling.}
    \item{In our experiments, we demonstrated that our upsampling model significantly reduces the performance gap in gait recognition tasks across LiDAR data with varying point cloud densities.}
\end{itemize}

\section{Related work}

\subsection{Gait recognition using LiDAR}
Traditional camera-based gait recognition methods can be broadly classified into two types: appearance-based approaches and model-based approaches.
The former focuses on extracting gait features directly from the visual appearances of the human body, such as images or videos \cite{gaitset, opengait}.
In contrast, the latter parameterizes visual data into human structures, such as shape-aware and non-shape-aware poses \cite{gaitgraph, smplgait}, and analyzes them to extract gait-related features.

Most existing studies on person identification using LiDAR sensors have employed appearance-based approaches with 2D representations, as the resolution of LiDAR sensors is generally lower than that of RGB cameras, making human pose estimation less effective.
Benedek \textit{et al.} \cite{benedek} proposed a projection-based model using gait energy images (GEIs) \cite{gei} to re-identify individuals in short-term scenarios. 
However, this method struggles to satisfactorily extract dynamic features from gait frames.
Yamada \textit{et al.} \cite{yamada} utilized temporal gait changes by employing long short-term memory (LSTM) networks with sequential range representations, optimized for processing efficiency based on the sensor's specifications. 
However, this approach is unsuitable for real-world scenarios, such as varying capture distances and pedestrian walking directions.
Ahn \textit{et al.} \cite{2v-gait-v1, 2v-gait-v2} explored view- and resolution-robust recognition frameworks by rearranging pedestrian point clouds based on a gait direction vector. 
Although this method enhances identification performance under complex confounding conditions, it still lacks the geometric features necessary for distance-independent analysis.
Shen \textit{et al.} \cite{lidargait} collected a large-scale LiDAR-based gait dataset, \textit{SUSTeck1K}, and designed a flexible and effective projection-based identifier. 
While this method shows promising results compared to camera-based methods \cite{opengait} at short measurement distances or with dense LiDAR sensors, such as Velodyne VLS-128, its performance degrades with longer distances or lower sensor resolutions, a challenge also noted by \cite{yamada}.
%

\subsection{Diffusion probabilistic models for LiDAR data generation}
Diffusion-based generative models \cite{ddpm, nonequilibrium, score_sde} have gained significant attention across a wide range of applications, including text-to-image generation, speech synthesis, translation, and compression.
Compared with generative adversarial networks (GANs) \cite{gan}, another prominent generative framework, diffusion models allow for stable training with a simple objective function by approximating likelihood maximization.
Specifically, denoising diffusion probabilistic models (DDPMs) \cite{ddpm}, a type of diffusion-based model, have demonstrated notably high fidelity for various completion tasks \cite{sr3, palette}.
These models learn the general distribution of a dataset by iteratively adding noise to the input data and then denoising it from Gaussian noise during the inference phase.

Several studies on LiDAR data tasks, primarily focusing on scene completion, have employed diffusion-based frameworks.
Zyrianov \textit{et al.} \cite{lidargen} utilized NCSNv2 \cite{ncsn-v2}, a score-based generative model, to train both the range and reflectance modalities using image representations.
Nakashima \textit{et al.} \cite{r2dm} adopted unconditional DDPMs, incorporating inpainting and timestep-agnostic techniques \cite{variational_diffusion_models, repaint}, to enhance both the fidelity and efficiency of LiDAR data synthesis for sim2real applications.
Sander \textit{et al.} \cite{sander} introduced LiDAR upsampling models based on conditional DDPMs \cite{palette}, achieving faster sampling while maintaining high fidelity compared with prior works \cite{r2dm, lidargen}.
In this work, we modify conditional DDPMs \cite{palette}, similar to the approach in \cite{sander}, for pedestrian point cloud completion, independent of the sensor distance.
\begin{figure*}[t]
    \centering
    \includegraphics[width=0.98\textwidth]{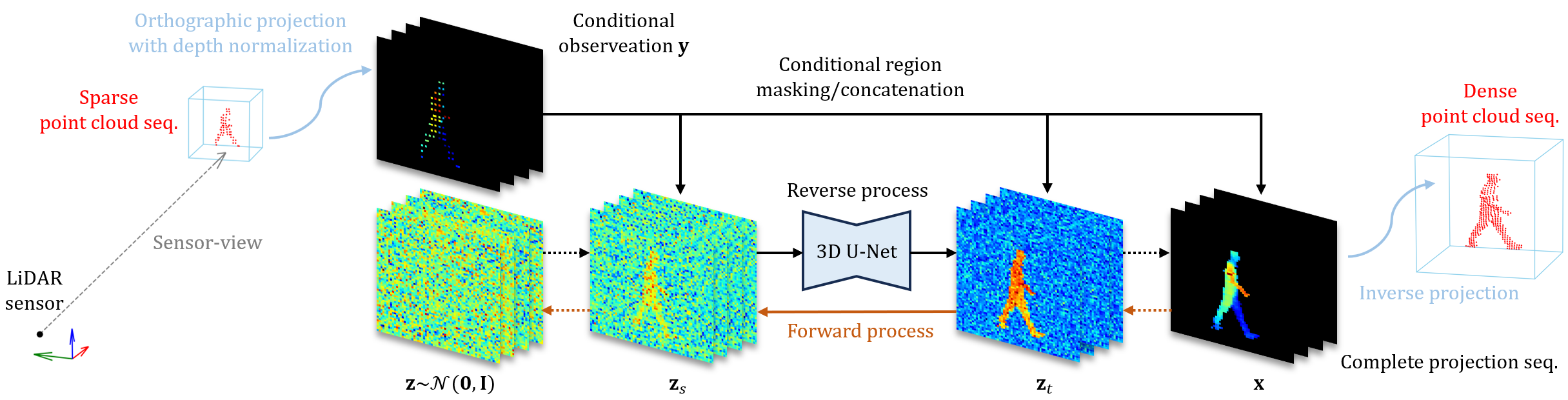}
    \caption{Overview of our upsampling pipeline. The diffusion processes operate within the orthographic projection domain with normalized depth values. The sampled depth projection sequences are then translated into 3D point cloud data.}
    \label{fig:network}
\end{figure*}
%

\section{Method}
In this section, we describe the problem of addressing missing parts in gait shape sequences and introduce the formulation of conditional DDPMs in relation to LiDAR data representation, loss function, and denoiser used for iterative refinement.

\subsection{Problem statement}
The 3D point cloud data captured from single LiDAR sensors can generally be transformed into range images.
Most existing studies \cite{sander, r2dm, lidargen} on LiDAR data generation have adopted a spherical projection function, which assigns an angular pixel to each angle: azimuth $\theta$ and elevation $\phi$.
This projection method provides well-aligned, one-to-one mapping and is cost-efficient for processing LiDAR data, as most LiDAR sensors used in autonomous driving are designed to spin mechanically and emit laser beams in a spherical pattern.
In contrast, orthographic projection directly maps LiDAR point clouds onto depth images within 3D Euclidean space.
Compared to spherical projection, the orthographic projection method \cite{2v-gait-v1, 2v-gait-v2, benedek} preserves the full size of objects regardless of varying measurement distances and does not rely on specific laser beam patterns from the sensors.
In addition, it eliminates the need for linear interpolation of pedestrian heights, which is often required in traditional camera-based gait recognition models \cite{opengait, lidargait}.
The missing points of the gait shapes in orthographic projection can be addressed using a distance- and emission-pattern-independent inpainting strategy, which is a type of linear inverse problem:
\begin{align}
    \mathbf{y} = \boldsymbol{H} \odot \mathbf{x}_0 + \epsilon,
\end{align}
where $\mathbf{x}_0$ represents a watertight gait video captured from a single LiDAR sensor, $\mathbf{y}$ is an incomplete gait video, $\boldsymbol{H}$ is a degradation noise mask, and $\epsilon$ represents noise.
In this context, we assume that the noise $\epsilon$ is set to zero.
Our goal is to solve this inpainting problem and recover $\mathbf{x}_0$ from the measurement $\mathbf{y}$ as a completed gait shape across varying measurement distances using conditional diffusion models.

\subsection{LiDAR data representation}
Based on the problem statement, we introduce an orthogonal projection method for LiDAR gait completion that can be directly visualized using sensors.
Given a pedestrian point cloud dataset $\mathcal{P}=\{\mathcal{P}^{j}_{i}|i=1,2,...,I; j=1,2,...,J_{i}\}$ with $I$ individuals and $J_{i}$ sequences for each individual $i$.
Each point cloud sequence $\mathcal{P}^{j}_{i} \in \mathbb{R}^{F \times N \times C}$ has $F$ frames, $N$ points for each frame $f$ and the number of channels $C$ represent Cartesian coordinates $(x, y, z)$.
Given a gait point cloud $\mathcal{P}^{j}_{i}$, we can define the center of mass $\textbf{c}^{j}_{i} = (c^{j}_{i, f, x}, c^{j}_{i, f, y}, c^{j}_{i, f, z})$ for frame $f$ as $\textbf{c}^{j}_{i} = \frac{1}{N} \sum^{N}_{n=1} \mathbf{p}^{j}_{i, f, n}$, where $c^{j}_{i, f, z}$ is set to zero because we only consider the sensors' emission directions on the $xy$-plane.
Subsequently, given a sensor-view angle ${\theta_{\text{sensor}}}^{j}_{i,f} = \arctan (c^{j}_{i, f, y}, c^{j}_{i, f, x})$ for the frame $f$ on the $xy$-plane for a given point cloud sequence $\mathcal{P}^{j}_{i}$, we can obtain the rotated point cloud sequence $\hat{\mathcal{P}}^{j}_{i} \in \mathbb{R}^{F \times N \times C}$ with a directional angle ${\theta_{\text{sensor}}}^{j}_{i,f}$ as follows:
\begin{align}
    \hat{\mathbf{p}}^{j}_{i, f, n} = (\mathbf{p}^{j}_{i, f, n} - \mathbf{c}^{j}_{i, f}) \cdot \mathbf{R}_{z}({\theta_{\text{sensor}}}^{j}_{i, f} + \pi), 
\end{align}
where $\mathbf{R}_{z}$ represents the rotation matrix around the $z$-axis.

As in \cite{2v-gait-v2}, we transform the point cloud sequence $\mathcal{P}^{j}_{i}$ into a gait image sequence $\mathbf{y}^{j}_{i} \in \mathbb{R}^{F \times 1 \times H \times W}$.
The gait image $\mathbf{y}^{j}_{i, f}$ of each frame $f$ has a resolution of $W(=l_y/r_y)$ in azimuth and $H(=l_z/r_z)$ in elevation and its depth value for an arbitrary point $\hat{\mathbf{p}}^{j}_{i, f, n}$ at each $(h, w)$ is determined as follows:
\begin{equation}
    h = \left\lfloor \frac{1}{r_{z}} \cdot ( \hat{p}^{j}_{i, f, n, z} - \underset{n \in \{ 1, ..., N \} }{ \text{min} } ( \hat{p}^{j}_{i, f=0, n, z} ) + l_{z-\text{const}} ) \right\rfloor,
\end{equation}
\begin{equation}
    w = \left\lfloor \frac{1}{r_{y}} \cdot ( \hat{p}^{j}_{i, f, n, y} + \frac{l_{y}}{2} ) \right\rfloor,
\end{equation}, 
where $l_z$ is the height of the $z$-axis, $l_y$ is the width of the $y$-axis, $r_z$ is the elevation resolution of $H$, $r_y$ is the azimuth resolution of $W$, and $l_{z-\text{const}}$ is the $z$-positional normalization constant for the generated gait video $\mathbf{y}^{j}_{i}$.
Here, when more than one point exists in the same pixel, the largest value is adopted, which is similar to the Z-buffer algorithm.
In this work, $l_z$, $l_y$, $r_z$, $r_y$, $l_{z-\text{const}}$, $H$, and $W$ are set to 2.6 m, 2.6 m, 0.04 m, 0.04 m, 0.3 m, 64, and 64, respectively.

\subsection{Preliminaries for DDPMs}
In this study, inspired by \cite{palette}, we build a diffusion-based inpainting model, as shown in Fig. \ref{fig:network}, conditioned on observation $\mathbf{y}$ (for simplicity, $j$ and $i$ are omitted).
Additionally, we employed the DDPM framework, which formulates transitions between data and latent spaces with continuous time $t \in [0, 1]$ \cite{variational_diffusion_models}.
Compared with discrete-time diffusion models \cite{ddpm}, a continuous noise schedule offers a finer approximation of the variational lower bound (VLB), leading to improved optimization efficiency.
In standard DDPMs, the process begins with Gaussian diffusion, where the data sample $\mathbf{x}_0$ is gradually corrupted by adding Gaussian noise from $t = 0$ (least noisy) to $t = 1$ (most noisy), resulting in a noisy version of $\mathbf{x}$, referred to as latent variable $\mathbf{z}_t$.

In the forward diffusion process, the distribution of latent variable $\mathbf{z}_t$ conditioned on $\mathbf{x}_0$ for any timestep $t$ can be given by:
\begin{equation}
    q(\mathbf{z}_t|\mathbf{x}_0) = \mathcal{N}(\alpha_{t} \mathbf{x}_0, \sigma^2_t \mathbf{I}),
\end{equation}
where $\alpha_t$ and $\sigma^2_t$ are strictly positive scalar-valued functions of $t$ in the noise schedule.
In this study, we employed $\alpha$-cosine schedule \cite{alpha-cosine}, which is one of the most popular schedules, resulting in $\alpha_{t} = \cos(\pi t / 2)$ and $\sigma_t = \sin (\pi t / 2)$.
Transition $\mathbf{z}_{t}$ can be tractably simplified using a re-parameterization trick as $\mathbf{z}_t = \alpha_t \mathbf{x}_0 + \sigma_t \epsilon$, where $\epsilon \sim \mathcal{N}(\mathbf{0}, \mathbf{I})$.
The signal-to-noise ratio of $\mathbf{z}_{t}$ can be defined as $\lambda_t = \alpha^2_t / \sigma^2_t$, where $\alpha_t = \sqrt{1 - \sigma^2_t}$ following the variance-preserving diffusion process \cite{score_sde}.
The transition of the latent variable $q(\mathbf{z}_t | \mathbf{z}_s)$ from timestep $s$ to $t$ for any $0 \leq s \leq t \leq 1$ is also Gaussian, written as:
\begin{equation}
    q(\mathbf{z}_t | \mathbf{z}_s) = \mathcal{N} (\alpha_{t|s} \mathbf{z}_s, \sigma^2_{t|s} \mathbf{I}), 
\end{equation}
where $\alpha_{t|s} = \alpha_t / \alpha_s$ and $\sigma^2_{t|s} = \sigma_t^2 - \alpha_{t|s}^2 \sigma_s^2$.
Given the above distributions, the reverse diffusion process $p(\mathbf{z}_s | \mathbf{z}_t)$ can be defined as: 
\begin{equation}
    p(\mathbf{z}_s | \mathbf{z}_t) = \mathcal{N} ( \mathbf{\mu}_t (\mathbf{x}_0, \mathbf{z}_t), \Sigma^2_t \mathbf{I} ), 
\end{equation}
where 
$\mathbf{\mu} (\mathbf{x}_0, \mathbf{z}_t) = \frac{\alpha_{t|s} \sigma^2_s}{\sigma^2_t} \mathbf{z}_t + \frac{\alpha_s \sigma^2_{t|s}}{\sigma^2_t} \mathbf{x}_0$ 
and 
$\Sigma^2_t = \frac{\sigma^2_{t|s} \sigma^2_s}{\sigma^2_t}$.

\subsection{Noise prediction model}
%
%
In this study, we used a modified 3D U-Net architecture \cite{vdm} as a noise prediction model $\hat{\epsilon}_\mathbf{\theta} ( \cdot )$, a parameterized neural network, to predict the noise $\epsilon$ and ensure consistent natural gait shapes across video frames $F$.
Compared with the standard U-Net architecture in \cite{ddpm, palette}, this 3D U-Net is factorized over space and time. 
Specifically, it includes space-only 3D convolution blocks, and the attention in each spatial attention block is applied over the space. 
In addition, temporal attention is used after each spatial attention block, with relative position embeddings applied in each temporal attention block, making it suitable for video data generation.

\begin{table*}[t]
    \centering
    \caption{Generative evaluation of the SUSTeck1K dataset with noise masks}
    \label{tab:generative_quality}
    \resizebox{1.0\linewidth}{!}{
        \begin{tabular}{ c c c | c c c c c c c c c}
        \hline
        \multicolumn{3}{c}{} & \multicolumn{9}{c}{Means (Test set)} \\
        \cmidrule(lr){4-12}
        \multicolumn{3}{c}{Upsampling} & \multicolumn{3}{c}{\textbf{V}$\times 1/2$, \textbf{P}$\times 1/6$} & \multicolumn{3}{c}{\textbf{V}$\times 2/3$, \textbf{P}$\times 2/6$} & \multicolumn{3}{c}{\textbf{V}$\times 3/4$, \textbf{P}$\times 3/6$} \\
        \cmidrule(lr){1-3} \cmidrule(lr){4-6} \cmidrule(lr){7-9} \cmidrule(lr){10-12}
        Approach & Method & Input Modality & PSNR $\uparrow$ & SSIM $\uparrow$ & Consistency $\downarrow$ & PSNR $\uparrow$ & SSIM $\uparrow$ & Consistency $\downarrow$ & PSNR $\uparrow$ & SSIM $\uparrow$ & Consistency $\downarrow$ \\ 
        \hline
        Interpolation & Nearest-neighbor & Depth Image & 6.90 & 0.031 & 0.041 & 6.84 & 0.029 & 0.043 & 6.78 & 0.025 & 0.045 \\
        Interpolation & Bilinear & Depth Image & 20.90 & 0.852 & 0.016 & 20.99 & 0.841 & 0.017 & 20.83 & 0.840 & 0.019 \\
        Interpolation & Bicubic & Depth Image & 21.05 & 0.855 & 0.017 & 21.08 & 0.843 & 0.017 & 20.90 & 0.842 & 0.019 \\
        \hline
        Diffusion & Palette \cite{palette} & Depth Image & 26.14 & 0.940 & 0.009 & 24.17 & 0.908 & 0.013 & 23.15 & 0.888 & 0.017 \\
        Diffusion & Ours w/o masking loss & Depth Video & 27.22 & 0.953 & \textbf{0.007} & 25.56 & \textbf{0.932} & \textbf{0.010} & 24.86 & \textbf{0.922} & \textbf{0.011} \\
        Diffusion & Ours & Depth Video & \textbf{27.27} & \textbf{0.954} & \textbf{0.007} & \textbf{25.59} & \textbf{0.932} & \textbf{0.010} & \textbf{24.89} & \textbf{0.922} & \textbf{0.011} \\
        \hline
        \end{tabular}
    }
\end{table*}
\begin{table*}[h]
    \centering
    \caption{Identification Evaluation using a LidarGait on SUSTeck1K dataset with noise masks}
    \label{tab:identification_susteck1k}
    \resizebox{1.0\linewidth}{!}
    {
        \begin{tabular}{ c c c | c c c c c c c c c }
        \hline
        \multicolumn{3}{c}{} & \multicolumn{9}{c}{Means (Probe set)} \\
        \cmidrule(lr){4-12}
        \multicolumn{3}{c}{Upsampling} & \multicolumn{3}{c}{\textbf{V}$\times 1/2$, \textbf{P}$\times 1/6$} & \multicolumn{3}{c}{\textbf{V}$\times 2/3$, \textbf{P}$\times 2/6$} & \multicolumn{3}{c}{\textbf{V}$\times 3/4$, \textbf{P}$\times 3/6$} \\
        \cmidrule(lr){1-3} \cmidrule(lr){4-6} \cmidrule(lr){7-9} \cmidrule(lr){10-12}
        Approach & Method & Input Modality & Rank1 $\uparrow$ & Rank5 $\uparrow$ & Rank10 $\uparrow$ & Rank1 $\uparrow$ & Rank5 $\uparrow$ & Rank10 $\uparrow$ & Rank1 $\uparrow$ & Rank5 $\uparrow$ & Rank10 $\uparrow$ \\
        \hline
         &  &  & 1.40 & 5.85 & 10.13 & 0.18 & 1.08 & 2.34 & 0.15 & 0.82 & 1.68 \\
        \hline
        Interpolation & Nearest-neighbor & Depth Image & 0.17 & 0.93 & 1.78 & 0.17 & 0.86 & 1.67 & 0.16 & 0.78 & 1.54 \\
        Interpolation & Bilinear & Depth Image & 1.35 & 5.16 & 8.52 & 0.62 & 2.58 & 4.86 & 0.44 & 1.96 & 3.72 \\
        Interpolation & Bicubic & Depth Image & 1.51 & 5.63 & 9.16 & 0.73 & 3.01 & 5.37 & 0.52 & 2.20 & 4.08 \\
        \hline
        Diffusion & Palette \cite{palette} & Depth Image & 23.62 & 48.69 & 61.07 & 9.93 & 26.61 & 37.31 & 7.16 & 13.79 & 21.82 \\
        Diffusion & Ours w/o masking loss & Depth Video & 31.69 & 58.57 & 70.27 & 18.07 & 40.72 & 53.08 & 11.38 & 29.72 & 41.16 \\
        Diffusion & Ours & Depth Video & \textbf{32.49} & \textbf{59.77} & \textbf{71.28} & \textbf{18.97} & \textbf{42.09} & \textbf{54.52} & \textbf{11.85} & \textbf{30.68} & \textbf{42.26} \\
        \hline
        \end{tabular}
    }
\end{table*}

\subsection{Loss function}
We define the objective function for our diffusion model to estimate unknown $\hat{\mathbf{x}}$ from latent variable $\mathbf{z}_t$ with a conditional observation $\mathbf{y}$. 
In this paper, the target of the loss function is set to the noise $\epsilon$, and the latent variable $\mathbf{z}_t$ for each timestep $t$ is repeatably initialized by combining the conditional masks $\mathbf{m} \in \mathbb{R}^{F \times 1 \times H \times W}$ according to observation $\mathbf{y}$ as follows:
\begin{align}
    \mathbf{z}_t \leftarrow \mathbf{m} \odot \mathbf{y} + ( \mathbf{1} - \mathbf{m} ) \odot \mathbf{z}_t, 
\end{align}
\begin{align}
    m_{(f, 1, h, w)} =
    \begin{cases}
        1, & \text{if} \quad y_{(f, 1, h, w)} > 0, \\
        0, & \text{otherwise}.
    \end{cases}
\end{align}
Subsequently, the loss function is defined by concatenating the observation $\mathbf{y}$ and initialized latent variable $\mathbf{z}_t$ along the channel axis, as follows:
\begin{align}
    \mathcal{L}_{T \rightarrow \infty} = \mathbb{E}_{\epsilon \sim \mathcal{N} (\mathbf{0}, \mathbf{I}), t 
 \sim \mathcal{U} ( 0, 1 )} [ || \hat{\epsilon} ( \text{concat} ( \mathbf{y} , \mathbf{z}_t ) ; \lambda_t ) - \epsilon ||^2_2 ].
\end{align}
After the training phase, the gait data can be sampled by recursively inferring $p(\mathbf{z}_s | \mathbf{z}_t)$, where $x$ is approximated by $\hat{\mathbf{x}}_\mathbf{\theta} = ( \mathbf{z}_t - \sigma_t \hat{\epsilon}_\mathbf{\theta} ( \text{concat} ( \mathbf{y} , \mathbf{z}_t ) ; \lambda_t ) ) / \alpha_t$ with a finite number of timesteps $T$ from $t=0$ to $t=1$.
In addition, we mask the loss to compute only the unknown regions in the depth videos for more efficient training, as in \cite{sander}.
%


\section{Experiments}
In this section, we demonstrate the effectiveness of our upsampling method on both the generation and gait recognition tasks.

\subsection{Datasets}
%
The performance of our model was evaluated using two datasets.
The first is the \textit{SUSTeck1K} dataset \cite{lidargait}, which is a well-known LiDAR point cloud benchmark for gait recognition.
It was collected using a 128-beam LiDAR scanner (Velodyne VLS-128), capable of capturing objects and surroundings with high resolution, allowing for dense measurements.
In addition, this dataset includes data from 1,050 identities, 12 gait attributes, and eight viewpoints, making it suitable for training both identification and generative models, as well as for evaluating general-purpose performance.
\begin{table}[t]
    \centering
    \caption{Comparison between two datasets}
    \label{tab: datasets}
    \resizebox{1.0\columnwidth}{!}
    {
        \begin{tabular}{c | c c c c | c c c}
        \hline
        Datasets & Sensors & Beams & V/H Resolutions & Subjects & Views & Distances \\
        \hline
        SUSTeck1K \cite{lidargait} & VLS-128 & 128 & 0.11$^\circ$/0.1$^\circ$ & 1,050 & 12 & 7.5 m \\
        Ours \cite{2v-gait-v2} & VLP-32C & 32 & 1.33$^\circ$/0.1$^\circ$ & 30 & 8 & 10, 20 m \\
        \hline
        \end{tabular}
    }
\end{table}
%

%
The second dataset \cite{2v-gait-v2} was collected using a 32-beam LiDAR scanner (Velodyne VLP-32C), which has a lower resolution (fewer vertical laser beams) than the sensor used in the \textit{SUSTeck1K} dataset, resulting in sparser pedestrian point clouds at the same measurement distances, as shown in Table \ref{tab: datasets}.
This dataset consists of 30 identities, 8 views, and 2 comparative distances, all captured with a single gait attribute (\textit{Normal}).
The rotation speed of the LiDAR sensor was the same for both datasets, operating at 10 frames per second (FPS).

\subsection{Implementation details}
\begin{figure}[h]
    \centering
    \includegraphics[width=0.48\textwidth]{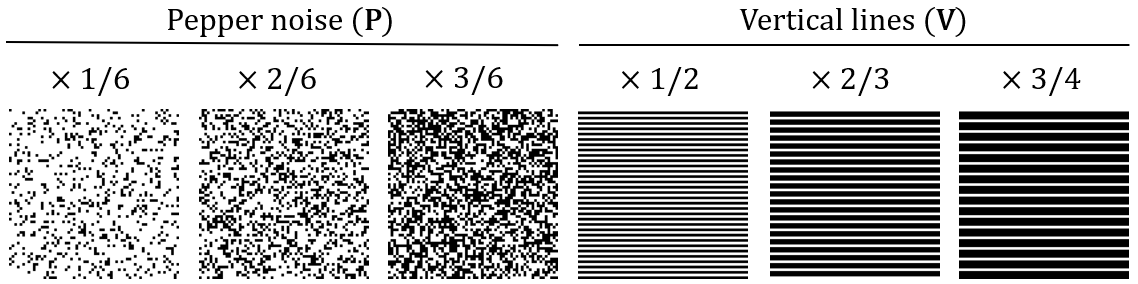}
    \caption{Noise masks used for training and testing our model. All mask sizes are $64 \times 64$, and the black regions in each binary noise mask indicate the points removed from the clean gait data.}
    \label{fig:masks}
\end{figure}

%
%
Following the original protocol, we used a subset of \textit{SUSTeck1K}, consisting of 250 subjects, for training our upsampling model.
We trained our model for 200,000 iterations with a learning rate of 0.0003 and 10 frames, while computing an exponential moving average (EMA) of the model weights with a decay rate of 0.995 every 10 steps.
We used two types of binary noise masks to degrade the original data during the training phase: pepper noise and vertical line masks, as shown in Fig. \ref{fig:masks}.
Pepper noise masks (\textbf{P}) are generated by randomly mapping points from a Bernoulli distribution to simulate noise in the azimuth based on the captured distances.
In contrast, the vertical line masks (\textbf{V}) represent the beam-level noise at the elevation of the LiDAR sensors.
In this study, we used three different ratios for each noise mask type and paired them with the original gait data during the training.
For the testing phase, we used the remaining test set of \textit{SUSTeck1K}, which consists of 800 subjects, randomly selecting 10 frames for each with three different combinations of noise masks.
In the applicability experiment, we used Ahn \textit{et al.}'s dataset \cite{2v-gait-v2}.
As a baseline, we compared our diffusion model with the vanilla Palette \cite{palette} using the proposed projection on two datasets.
In addition, we compared the well-established linear interpolation methods: Nearest-neighbor, Bilinear, and Bicubic.
In the generative quality evaluation, we adopted the following two widely-used standard metrics: the Peak Signal-to-Noise Ratio (PSNR), and Structural Similarity Index (SSIM).
We also focused on Consistency, which is a metric for video generation that evaluates temporal coherence by calculating the gradient between consecutive video frames.
In the gait recognition task, we used LidarGait \cite{lidargait} as the representative state-of-the-art identification model, which was pre-trained on the training set of \textit{SUSTeck1K}, following the original protocol.
In this study, identification accuracy refers to the average of the results obtained from all cross-views and gait attributes.
In addition, all gallery sets consisted of clean data, whereas noise masks were applied to the probe sets during testing on the \textit{SUSTeck1K} dataset.
All diffusion-based models were configured with a fixed timestep $T$ of 32.
For this study, both our upsampling model and LidarGait, trained on the training set of \textit{SUSTeck1K}, were applied in all experiments.
\begin{figure}[tb]
    \centering
    \includegraphics[width=0.48\textwidth]{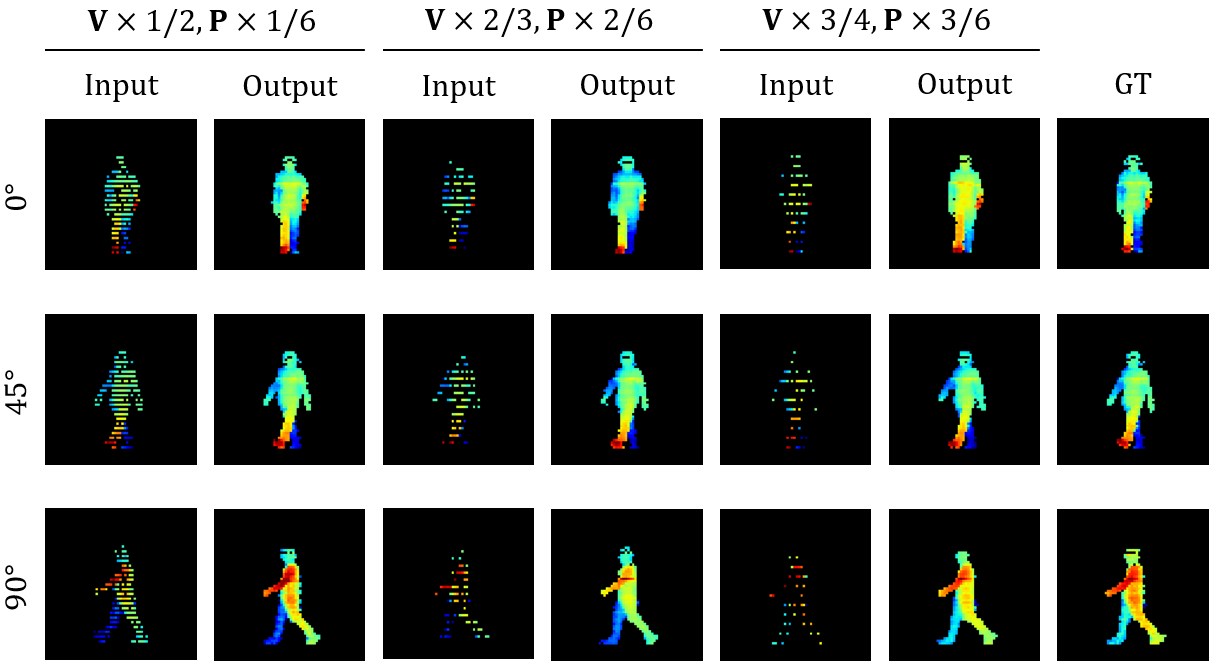}
    \caption{Upsampled results using our model on the \textit{SUSTeck1K} dataset for the \textit{Normal} attribute with three noise mask combinations. The redder the color, the greater the depth value.}
    \label{fig: angles}
\end{figure}
\subsection{Generative evaluation}

\begin{figure*}[t]
    \centering
    \includegraphics[width=0.98\textwidth]{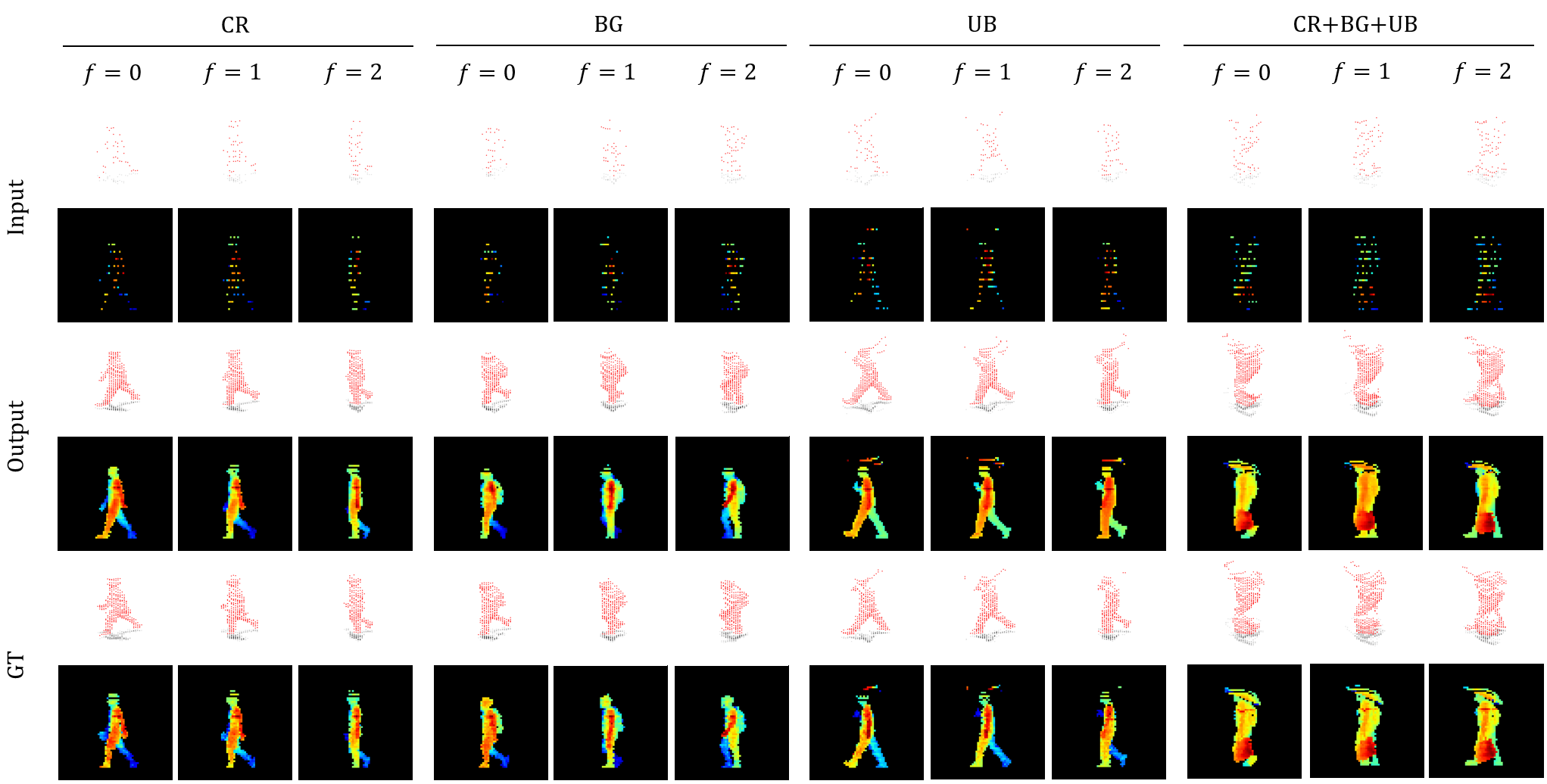}
    \caption{Upsampled results using our model from noise masks with \textbf{V}$\times 3/4$ and \textbf{P}$\times 3/6$. We showcase the samples for three gait variances: Carrying (\textit{CR}), Bag (\textit{BG}), and Umbrella (\textit{UB}).}
    \label{fig: attributes}
\end{figure*}

The quantitative generative evaluation results on the \textit{SUSTeck1K} dataset are listed in Table \ref{tab:generative_quality}, and the examples sampled by our model for the gait attribute \textit{Normal} are shown in Fig. \ref{fig: angles}.
In Table \ref{tab:generative_quality}, diffusion-based methods significantly outperformed the interpolation approach across all three metrics.
Comparing our model to Palette \cite{palette}, we observed that our video-based model \cite{vdm} is more effective than the image-based approach.
Notably, as the noise masks became more severe, the performance gap between our model and Palette increased.
For another gait attributes in Fig. \ref{fig: attributes}, we can see that our method realizes high fidelity in both 2D projected gait shapes and the structure of 3D point clouds.

\subsection{Gait recognition task}
\begin{figure}[h]
    \centering
    \includegraphics[width=0.48\textwidth]{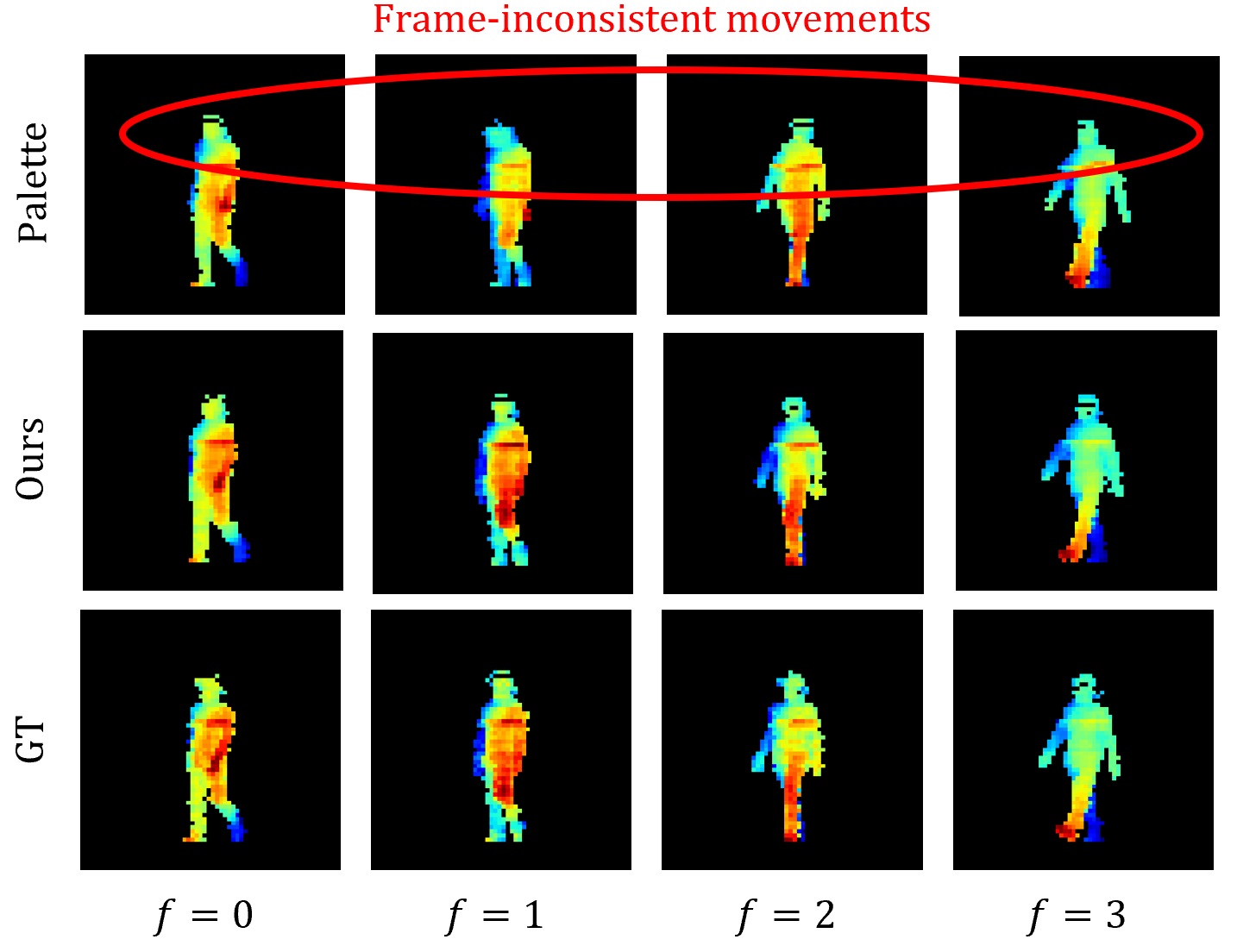}
    \caption{Comparison between our model and Palette \cite{palette}. The results are sampled from noise masks with \textbf{V}$\times 3/4$ and \textbf{P}$\times 3/6$ (top two rows).}
    \label{fig:palette}
\end{figure}

\begin{figure}[h]
    \centering
    \includegraphics[width=0.48\textwidth]{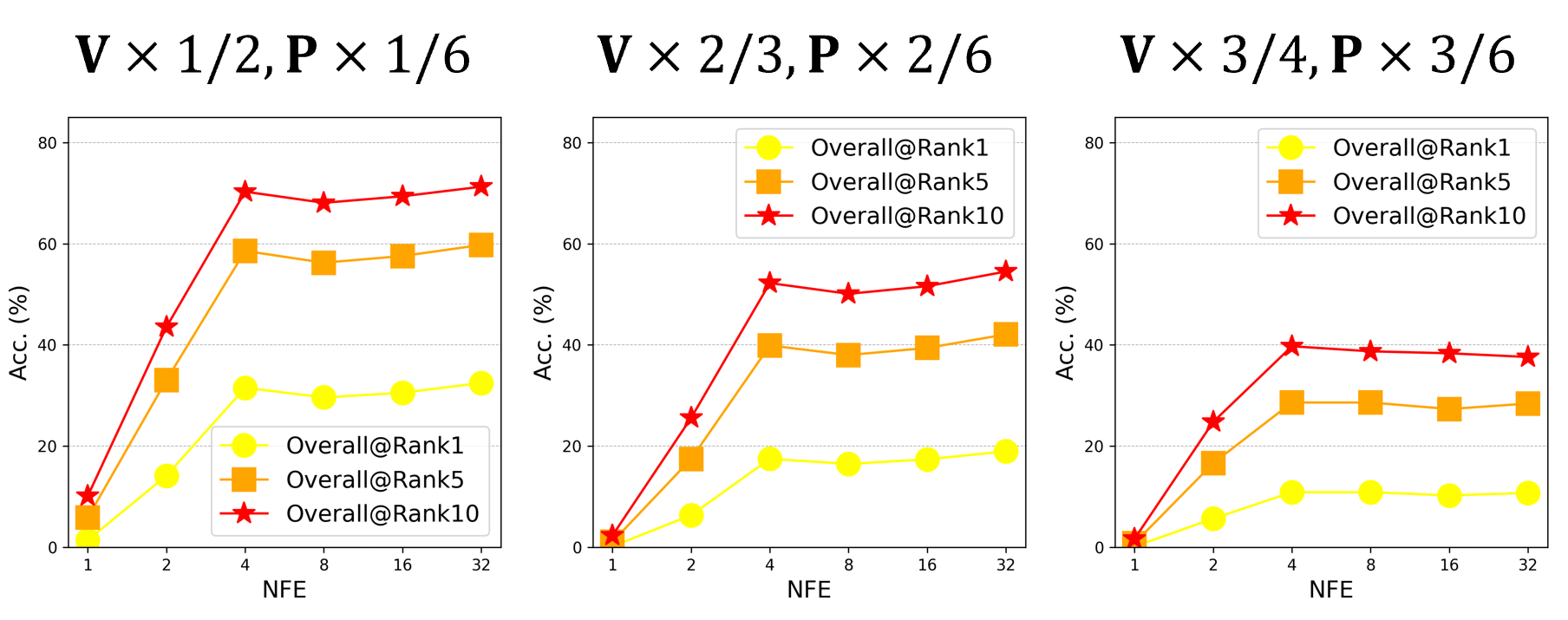}
    \caption{Comparison of the number of function evaluations (NFE) for our model by sweeping $T$ across $\{1, 2, 4, 8, 16, 32\}$.}
    \label{fig:nfe}
\end{figure}

%
The identification results conducted on \textit{SUSTeck1K} using LidarGait \cite{lidargait} are listed in Table \ref{tab:identification_susteck1k}.
In Table \ref{tab:identification_susteck1k}, we can see that the interpolation approach achieves little to no improvement in recognition performance on sparse gait data.
Similar to Table \ref{tab:generative_quality}, it can be observed that our model outperforms Palette as the noise level in the probe set increases because of its ability to maintain consistency in appearance across gait sequences, as shown in Fig. \ref{fig:palette}. 
Fig. \ref{fig:nfe} shows the identification evaluation scores as functions of the number of function evaluations (NFE), which indicate how many times the neural networks are processed during sampling.
For all noise mask combinations, it can be observed that overall performance generally improves as $T$ increases, while remaining consistent even when $T$ is reduced to 4.
\subsection{Application}

\begin{figure}[h]
    \centering
    \includegraphics[width=0.48\textwidth]{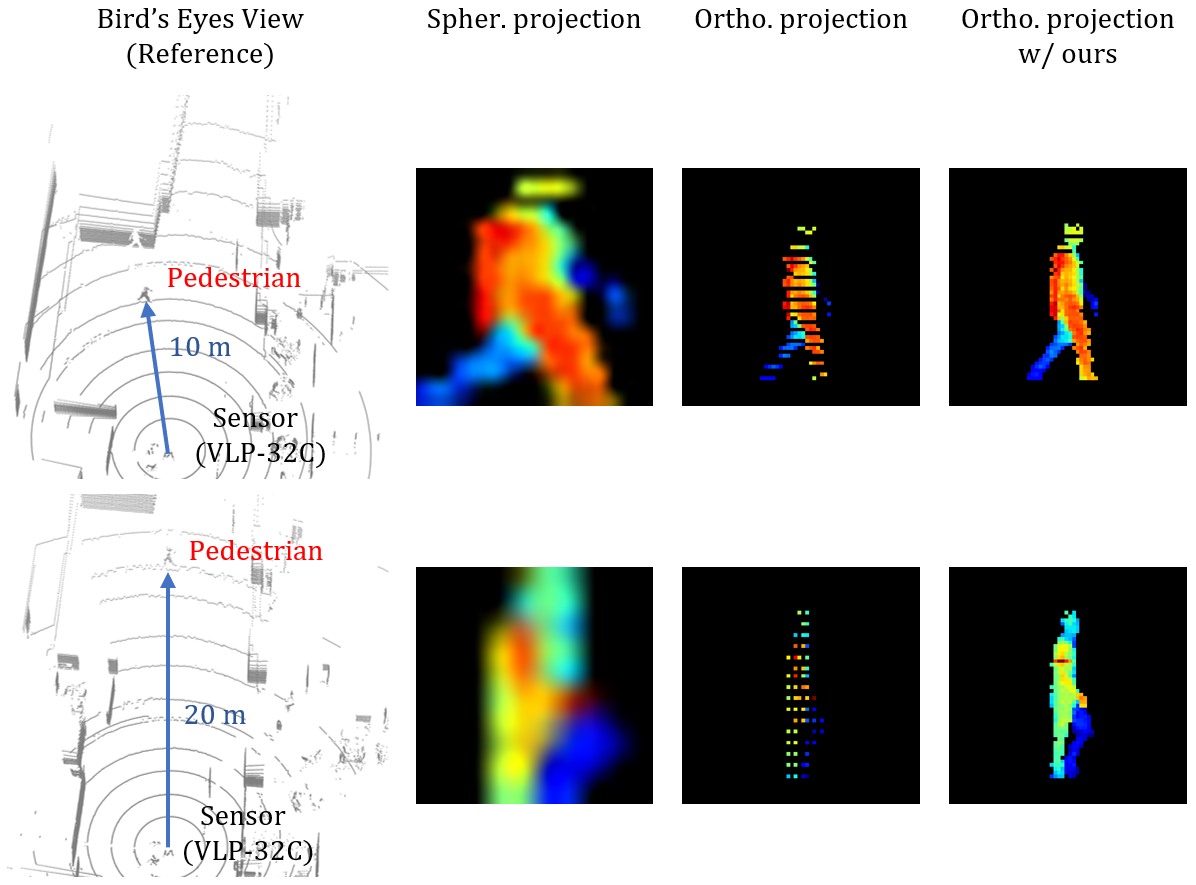}
    \caption{Projection comparison on the real-world dataset \cite{2v-gait-v2} at two capture distances: spherical projection (Spher.) and our orthographic projection (Ortho.).}
    \label{fig:kugait}
\end{figure}

The identification results conducted on our dataset \cite{2v-gait-v2} using LidarGait \cite{lidargait} to evaluate the applicability of our model are shown in Table \ref{tab:identification_kugait}. 
In addition, the gait shapes according to the two different point cloud projections as illustrated in Fig. \ref{fig:kugait}.
In Table \ref{tab:identification_kugait}, we observed that our upsampling method and training strategy significantly contribute to performance improvement, even for real-world scenarios.
Interestingly, the highest performance gain was achieved when both the probe set and the gallery set were fully restored.
\begin{table}[h]
    \centering
    \caption{Identification results on the real-world dataset \cite{2v-gait-v2}.}
    \label{tab:identification_kugait}
    \resizebox{1.0\columnwidth}{!}
    {
        \begin{tabular}{ c c c c | c c }
        \hline
        \multicolumn{3}{c}{Upsampling} & \multicolumn{1}{c}{} & \multicolumn{2}{c}{Overall} \\
        \cmidrule(lr){1-3} \cmidrule(lr){5-6}
        Method & Gallery (10 m) & Probe (20 m) & Projection & Rank1 $\uparrow$ & Rank5 $\uparrow$ \\
        \hline 
         & & & Spher. & 5.51 & 25.98 \\
        & & & Ortho. & 7.07 & 30.80 \\
        \hline
        Palette \cite{palette} & & \checkmark & Ortho. & 19.57 & 56.25 \\
         & \checkmark & \checkmark & Ortho. & 25.45 & 63.54 \\
        \hline 
        Ours & & \checkmark & Ortho. & 21.28 & 60.94 \\
         & \checkmark & \checkmark & Ortho. & \textbf{25.97} & \textbf{66.82} \\
        \hline
        \end{tabular}
    }
\end{table}
%

\section{Conclusions}
In this study, we introduced a diffusion-based upsampling method for LiDAR-based gait sequence data, addressing a distance-independent inpainting problem.
Our model demonstrated significant performance improvements compared with other methods in terms of both generation quality and gait recognition.
Notably, our model is effective even for pedestrians with varying sensor resolutions or measurement distances in real-world scenarios.
Future work will involve applying point-based identification models and investigating restoration for additional noise types, such as frame-drop noise and occlusions caused by obstacles.
\addtolength{\textheight}{-12cm}   
\bibliographystyle{plain}
\bibliography{ref}

\end{document}